\documentclass[pdflatex,sn-mathphys-num]{xx}
\usepackage{graphicx}%
\usepackage{multirow}%
\usepackage{amsmath,amssymb,amsfonts}%
\usepackage{amsthm}%
\usepackage{mathrsfs}%
\usepackage[title]{appendix}%
\usepackage{xcolor}%
\usepackage{textcomp}%
\usepackage{manyfoot}%
\usepackage{booktabs}%
\usepackage{algorithm}%
\usepackage{algorithmicx}%
\usepackage{algpseudocode}%
\usepackage{listings}%
\usepackage{amsmath,booktabs}
\usepackage{multirow}
\theoremstyle{thmstyleone}%

\usepackage{array}
\theoremstyle{thmstyletwo}%
\usepackage{tikz}
\theoremstyle{thmstylethree}%
\usepackage{hhline}
\usepackage{tcolorbox} 
\newcommand{\xtiny}{\fontsize{6}{8}\selectfont}
\usepackage{adjustbox}
\usepackage{tabularx}
\usepackage{xcolor}  
\usepackage{pgfplots}
\definecolor{xblue}{RGB}{66,133,244}

\definecolor{xred}{RGB}{234,67,53}
\usepackage{float}
\definecolor{xyellow}{RGB}{251,188,5}

\begin{document}
\title[Article Title]{Collaborative and Proactive Management of Task-Oriented Conversations}
\author[1]{\fnm{Arezoo} \sur{Saedi}\footnotemark} \email{a.saedi@eng.ui.ac.ir  }

\footnotetext{Ph.D. candidate at the University of Isfahan, Visiting researcher at Paris-Saclay University}
\author*[1]{\fnm{Afsaneh} \sur{Fatemi}}\email{a\_fatemi@eng.ui.ac.ir}

\author[1]{\fnm{Mohammad Ali} \sur{Nematbakhsh}}\email{mnematbakhsh@eng.ui.ac.ir}

\author[2]{\fnm{Sophie} \sur{Rosset}}\email{sophie.rosset@lisn.fr}
\author[2]{\fnm{Anne} \sur{Vilnat}}\email{anne.vilnat@universite-paris-saclay.fr}

\affil*[1]{\orgdiv{Faculty of Computer Engineering}, \orgname{University of Isfahan}, \orgaddress{\city{Isfahan}, \country{Iran}}}

\affil[2]{\orgdiv{LISN}, \orgname{Université Paris-Saclay, CNRS}, \orgaddress{\city{Orsay}, \country{France}}}

\abstract{Task-oriented dialogue system (TOD) completes particular task based on user preferences across natural language interactions. Considering the impressive performance of large language models (LLMs) in natural language processing (NLP) tasks, most of the latest TODs are centered on LLMs. While proactive planning is crucial for task completion—the primary purpose of TODs— many existing TODs overlook effective goal-aware planning, partly due to deficient incorporating of intermediate conversation information essential for informed decision-making. This paper presents a model for managing task-oriented conversations, conceptualized centered on the information state approach to dialogue management. The created model incorporates constructive intermediate information in planing. Initially, predefined slots and text\_part informational components are created to model user preferences. Consequently, while investigating task-oriented conversations, critical circumstances in interactions are identified, and informational components corresponding to these circumstances are created. Possible configurations for these informational components leads to limited information states. Then dialogue moves, which indicate movement between these information states and the procedures that must be performed in this movement, are created. Exploiting these information states and dialogue moves, the update strategy as a pivotal part of presented model is constructed which indicates the dialogue moves in conversation. Leveraging in-context learning capabilities of LLMs, the created model is implemented. In this model, database queries are created centered on indicated predefined slots and the order of retrieved entities is indicated centered on the text\_part. This mechanism enables passing the whole corresponding entities to the user preferences in the order of congruency. Evaluations exploiting the complete test conversations of MultiWOZ 2.2, with no more than a domain in a conversation, illustrate maximal inform, and success. Compared with previous methods, the performance of the created domain-independent model improves in terms of these task completion metrics.}
\maketitle

\section{Introduction}\label{xintro}
The primary purpose of a collaborative task-oriented dialogue system (TOD) is to manage conversation and coach a user to complete task without any conflict of interest with the user. The collaborative TOD leverages its whole potential to effectively complete the task centered on user preferences. Management of conversations within TOD involves updating the context centered on previous interactions, interpreting user utterances within the context, interfacing with back-end infrastructure such as databases, planners, and execution modules pertinent to the current task or domain, and creating the later utterance \cite{traum2003information, speechlanguageprocessing,deng2023prompting}. \\
Predominantly, many of the latest task-oriented dialogue systems (TODs) are created centered on the dialogue state model. This frame-based model characterizes slots as markers of particular pieces of predefined information that TODs intend to extract from user utterances in the course of the interaction. The dialogue state model comprises natural language understanding (NLU), dialogue state tracker (DST), dialogue policy (DP), and natural language generation (NLG) components.\cite{speechlanguageprocessing}.\\
The considerable portion of processing in TODs is language processing. With the prowess of the large language model (LLM) in mitigating linguistic challenges—particularly contextual understanding and language generation—LLMs are incorporated in the majority of lately introduced TODs. In task-oriented conversations, despite chitchat, the purpose is task completion, with each TOD-created utterance in line with this end. While many previous LLM-based TODs fail to maintain this goal-oriented perspective, proactivity is a crucial prerequisite for creating a goal-aware TOD. Employing the information of intermediate phases in the conversation, proactivity enables the TOD to consider the consequence of changes in task completion and provide effective unprompted information ; to illustrate, if there is no entity in the database that is congruent with user-indicated configuration of predefined slots, a proactive TOD, considering the current configuration and entities in the database, can create a clarifying question which enables the user to change preferences slightly and complete the task. While different methods were employed in previous works, the intermediate information they could integrate into the planning process is commonly limited to the dialogue state, acts and database-retrieved information. The lack of comprehensive constructive intermediate information in dialogue state is a limitation of previous TODs and prevents proactivity\cite{deng2023prompting,deng2023survey,deng2023rethinking}.\\
The information state-based dialogue theory framework comprises informational components, their formal representations, dialogue moves, update rules, and update strategies. Contrasting with the dialogue-state model, the information-state approach to dialogue management presents a wider perspective on modeling task-oriented conversations. The dialogue-state model can be conceptualized as an instance of the information-state model. In addition, while the dialogue-state model is in contrast to plan-based methodologies, the information-state model enables an effective integration of these paradigms. \cite{traum2003information}.\\
In this paper, the challenges and defeats in the conversations of previous TODs investigated to identify which missing intermediate information leads to error-prone conditions. Instances of these challenging conditions might include user input errors in predefined slots or conditions in which users reject the entities. Then, centered on this investigation and inspired by the conceptualization of information state approach to dialogue management, limited informational components corresponding to constructive intermediate pieces of information in task-oriented conversations are created. Initially, considering the limitation of creating a TOD centered on limited predefined intents \cite{bocklisch2024task}, for modeling user preferences, informational components of predefined slots and text\_part are created, while the text\_part contains the constraint in preferences beyond the predefined slots. Then, challenging conditions in conversations examined, and constructive intermediate information, which employing it could mitigate that challenge and informational components corresponding to them are created. The configurations of created informational components leads to limited information states. \\
In order to construct a proactive TOD, not only creating informational components are important but also constructing an effective management policy is crucial. Consequently, dialogue moves corresponding to the possible movement between information states are initiated and their corresponding procedures, including language processing, database retrieval and changing informational component configurations that will be performed in these movements, are established. Then centered on the possible information states and created dialogue moves, the update strategy is constructed to manage changes in conversation, indicating a dialogue move in each phase.\\
In the model presented in this paper, the task of TOD is decomposed into smaller parts centered on critical points in the conversation, which contain constructive pieces of intermediate information. Pre-training an LLM or fine-tuning a pre-trained LLM are popular manners to create LLM-based TODs. To implement modularized TODs models in this way, the cost of curating datasets and trainings is not negligible. With considerable progress in instruction-tuned LLMs, leveraging in-context learning has become common to employ the capabilities of LLMs out-of-the-box in conducting particular tasks. While writing the effective prompt that clearly indicates the instruction is profound for in-context learning of LLMs, without updating LLM parameters or imposing curating datasets, the cost of learning is minimal \cite{dong2022survey,deng2023prompting}. \\
The language proficiency of models centered on in-context learning of LLM is considerable, but effective planning is commonly ignored in these models. Therefore, methods including task decomposition and external planner-aided are presented to improve the planning of these LLM-based models. \cite{huang2024understanding}. The presented model is implemented centered on in-context learning of LLMs inspired by task decomposition and external planner-aided methods. As in-context learning enables injecting intermediate information into established language processing subtasks and performing them with low cost, it can be potentially infeasible if the implementation centered on pre-training or fine-tuning LLMs.\\
The implemented method can be interpreted as a retrieval-augmented generation (RAG) model, while the retrieval component is not centered on the vector database but on a custom entity search model. The prompt in in-context learning is created by integrating the concluded information of retrieval and the informational components into the prompt template. In the presented model indicated in \ref{xpresentedmodel}, the update strategy, considering information state, infers dialogue move. Procedures corresponding to dialogue move may include language processing, retrieving entity information from the database, or configuring informational components. After performing procedures, its product is passed to the update strategy. If the ToD utterance is prepared, it is presented to user; if it is not created, the update strategy indicates dialogue move considering the information state, then the procedures corresponding to that dialogue move are performed, and its product is passed to update strategy. If the ToD utterance is prepared, it is presented to the user. If the conversation is not completed, the user passes another utterance to the model. The update strategy pursues processing. This cycle continues until the conversation is ended. \\
\begin{figure}[H]
    \includegraphics[width=\textwidth]{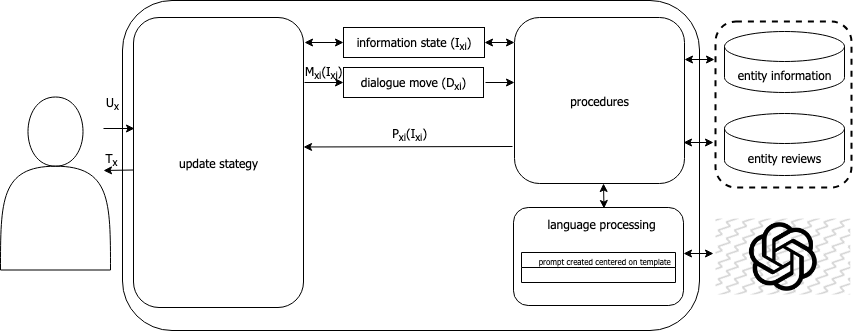}
    \caption{Created Model}
    \label{xxpapermodel}
\end{figure}
The performance of implemented TOD, centered on task completion metrics, is evaluated and compared with previously created TODs in the collection of MultiWOZ conversations, while each conversation is not in more than a domain. The inform of the presented method is complete and considering exploitation of the custom retrieval mechanism, there is no drop in success compared with inform, which is confirmed in evaluations. The presented method improves the inform and susses of previous models considerably.\\
The purpose of this paper is to improve task completion in task-oriented dialogue systems through collaborative proactive planning, which integrates comprehensive intermediate information into the dialogue management. These tasks commonly lead to retrieving entity information from database. This paper creates a collaborative model for management of task-oriented conversations that:
\begin{itemize}
    \item Incorporates comprehensive intermediate information crucial for planning, enhancing task completion and reducing error-prone conditions.
    \item Leverages the language proficiency of large language models (LLMs) while is proactive and task-aware in its planning.
    \item It is centered on the information-state approach to dialogue management, which enables it to mitigate the limitations of dialogue-state-based models.
    \item It is centered on in-context learning of LLMs, inspired by task decomposition and planner-aided methods to improve LLMs planning capabilities.
    \item It exploits an entity search mechanism that, after retrieving entities centered on predefined slots, orders entities considering their congruence with the rest of preferences.
\end{itemize}
In section \ref{Previous Works}, previous works and their comprehensive investigation are provided. Ensuing this, section \ref{Method} introduces the presented model. A comprehensive breakdown of implementation, experiments and evaluations are then presented in section \ref{Experiments}. Eventually, conclusions are provided in section \ref{Conclusion}.
\section{Previous Works}
\label{Previous Works}
The informed management of interaction in TOD is a critical information science issue \citep{info}. Previous works introduce different models to create TODs. With the capabilities of large language models and considerable language processing within TODs, most of the lastly implemented TODs are LLM-based. The previous works in LLM-based TODs in \ref{premethods}, and papers concentrating on proactivity and creating clarifying questions in \ref{proactive} are explored. The pivotal insights and challenges in previous works for creating proactive collaborative TOD are investigated in \ref{xDemerits}. 
\subsection{LLM-based models }
\label{premethods}
Initially, the models centered on pre-training or calibrating LLMs are investigated in \ref{premethodstraditional}. Then, the models centered on in-context learning of LLMs, which exploit LLMs out-of-the-box without changing their parameters \citep{dong2022survey}, are examined in \ref{in_context_learning}. 
\subsubsection{Models Centered on Pre-training and Fine-tuning of LLMs  }
\label{premethodstraditional}
 \citep{peng2021soloist} endeavors to mitigate the expenses of creating TODs on a large scale and introduce a benchmark for improving them. The costs incurred in creating TODs include labeling and providing domain-specific information and expertise. While language models exhibit considerable prowess in creating utterances, their calibration for crafting utterances tailored to the goals and domain within TODs is imperative. Consequently, their paper presents an loss function encompassing belief state tracking and response generation subtasks.\\
Exploiting their created loss, they create and train an LLM-based model for task-centered utterance creation. Their model emerges as a pre-trained language model tailored explicitly for TODs. Customization of their presented model for individual domains involves calibrating the model centered on created loss function, exploiting pre-collected in-domain conversations. The process of collecting data for each domain is expensive. To mitigate this challenge, their paper presents a multi-phase calibration method. Ensuing initial calibration with a pre-collected dataset, they continue improving the model exploiting a machine teaching method until either performance plateaus or certain indicated epochs have been completed.\\
TODs consist of modules for understanding, dialogue management, and utterance creation. While, the conventional modular approach in constructing such systems commonly introduces propagation errors. \citep{hosseini2020simple} presents a method to model the sequence probability, incorporating outputs from understanding, dialogue management, and utterance creation phases leveraging the chain rule of probability. They consequently train an LLM-based end-to-end model, leading to a unified language model for the whole modules. Their model improves its prior works on the multiWOZ dataset.\\
While complexity of database queries and utterances with conversation capabilities presents obstacles in pre-training an end-to-end TOD, the scarcity of training data exacerbates these challenges. \citep{qin2023modularized}  presents a model that decomposes the dialogue system into distinct knowledge base retrieval and generation components, which are pre-trained independently. This method enables exploitation of KB-free corpora in training the utterance creation component. In addition, their paper introduces a data augmentation method to mitigate the scarcity of data for training the database retrieval component, which enables effective calibration of their presented TOD.\\
While many previous methods to pre-training TODs have primarily emphasized understanding and creating utterances, \citep{he2022galaxy} concentrates on enhancing dialogue policy to improve pre-training language models in creating a TOD. They introduce a model centered on UniLM \citep{dong2019unified} for pre-training of TODs, which predicts the dialogue policy explicitly from labeled and unlabeled date, exploiting semi-supervised method. They create UniDA, which is labeled with dialogue acts, and create an unlabeled UnDial corpus, which is created through collecting certain dialogue corpora. The UniDA dataset is meticulously crafted from online dialogue corpora, and to ensure quality, they present a gating mechanism to choose only high-quality data points in pre-training.\\
Their presented model encompasses multiple objectives: response selection, response generation, dialogue action prediction, and consistency regularization. Their pre-training dataset is the union of UniDA and UnDial. Their presented objective for labeled data is the cumulation of the response selection, response generation, dialogue action prediction, and consistency regularization terms, while their objective for unlabeled data is the cumulation of: response selection, response generation with multiply of consistency regularization and gating score. Their presented objective for calibration depends on whether the training dataset is labeled or unlabeled. For a labeled dataset, it is the cumulation of response selection, response generation, dialogue action prediction terms, while for unlabeled datasets it only consists of response selection, and response generation terms.\\
Pre-trained language models (PLMs) enhance the performance of TODs considerably. While the devoid of grounding and lack of diversity are issues within TODs created centered on pre-trained language models. paper \citep{kulhanek2021augpt} presents training objectives for calibrating PLMs to mitigate these challenges. Their presented model creates utterances, predicts beliefs, and preforms multiple subtasks.\\
While the parameters of belief state predictor and response predictor are common, the training loss for belief prediction and utterance prediction differ slightly. Their presented objective for belief prediction is centered on cross-entropy. While for utterance prediction, the objective includes cross-entropy and unlikelihood loss terms, to enhance diversity by penalizing repetition. They exploit a back-translation method to create paraphrases, thereby overcoming the lack of diversity obstacle.\\
\citep{su2022multi} concentrates on the inter-dependency of components within PLM-based TODs, including error propagation, data annotation overhead, and inference latency challenges. To mitigate these issues, they present a unified model for whole subtasks in task-oriented conversations, encompassing dialogue state tracking, dialogue policy learning, and natural language understanding and creation. Their paper introduces a pre-training model centered on a maximum likelihood objective to predict intended text centered on the context and a task-specific prompt. They indicate that exploiting this prompt inspired by the concept of in-context learning.\\
\citep{yang2021ubar} investigates utterance-level training and evaluation in previously presented methods to create end-to-end TODs centered on PLMs, which cause not only missing of the intermediate information, such as belief states, but also unrealistically  center on independent labeled information for subtasks. They model task-oriented conversations in conversation level and present a method centered on calibrating a large PLM on the whole context encompassing user utterance, belief state, database results, system act, and system utterance. In their presented method, calibration leverages the language modeling objective to optimize the probability of predicting the next word, and they probe the evaluation centered on the created content instead of predefined correct information.\\
While task-bot is a conversational task assistant (CTA) which equips users with information and enables them to complete the task independently, and is a little different from TOD, the components of them are not quite different \citep{gottardi2022alexa}. \citep{Tech2023} presents a task-bot model that in its dialogue manager, after identifying the task, retrieves the information from the corresponding databases. Exploiting this information, it updates PlanGraph which introduced in their paper and contributes to effectively managing complex tasks within task-bots. The embedded information in PlanGraph enables providing instructions. Throughout this process, certain filters are exploited to eliminate unsafe and harmful instructions. The dialogue manager identifies whether a template-based or neural creator, which is a question-answering module, will be exploited to create the utterance to maintain a balance between precision and flexibility. They create question-answering (QA) module through calibrating a pre-trained QA language model.
\subsubsection{Models Centered on In-Context Learning of LLMs }
\label{in_context_learning}
The critical role of language understanding and utterance creation tasks in task-oriented dialogue management, as well as the effectiveness of LLMs in performing these tasks, led to the creation of many LLM-based task-oriented dialogue managers. InstructTODS \citep{chung-etal-2023-instructtods} is an end-to-end TOD that leverages the capabilities of a large pre-trained language model out-of-the-box. While it withholds calibrating the exploited LLM, and eschews exploiting domain information or otologies, its performance in task completion competes that of calibrated TODs. The presented model of InstructTODS incorporates proxy belief states and action thoughts, enhancing its capability to interact effectively with the knowledge base. The LLM in utterance generation module is capable of incorporating not only the context but also entire database in its input. The proxy belief state encapsulates the user's intent until the current point in the conversation.Action thoughts are constructed centered on information encapsulated in the proxy belief state. Indeed, action thoughts work as indicators of the information that must retrieved from the knowledge base. Utilizing this information and the predefined templates, InstructTODS exploits the large language model as a code creator to create queries, exploiting action thoughts as input. The retrieved information is investigated upon querying and parsed by LLM to wrap-up the interaction and indicate whether the action thought has been completed. If not, another action thought is created, and the explained phases are iterated until the action thought is either completed or the knowledge retriever indicates a lack of corresponding information in the knowledge base. Then, the utterance creation module exploits the context and the concluded information of the knowledge base query process to create the utterance. In conditions which corresponding information is not in the knowledge base, the created utterance prompts the user for information.\\
\citep{gao-etal-2023-adaptive} presents a method for creating prompts in TODs. While leveraging LLMs in a box manner for TOD creation is practical, prompt creation for these systems tackles certain challenges. If creators construct a unified prompt to create utterances in the whole conversation turns, they must encompass constraints corresponding to every conversation condition within this prompt, inevitably leading to lengthy prompts which probably cause weak performance and hallucinations in creating utterances. While if creators construct a compact prompt, it might leads to oversight of crucial constraints corresponding with certain conversation conditions, undermining the TOD's proficiency in managing conversation. To mitigate this problem, \citep{gao-etal-2023-adaptive} introduces a method to dynamically create prompts at each conversation turn, considering the conversation context leading up to that point.\\
In their presented method, a trainable slot generator component is incorporated, which leverages context to identify the domain and slots within the belief state. The domain and slot information, extracted through this component, are considered prior information. Moreover, an adaptive prompt generator component exists within their presented model. Leveraging prior information and context, this component constructs prompts to enable the creation of utterances and the complete belief state. The prompts are constructed to create the complete belief state, comprising constant and changing elements. The constant portion encompasses instructions for belief state extraction along with illustrative examples, whereas the changing portion encompasses domain slots and their corresponding candidate configurations. The prompts constructed for creating utterances also comprise constant and changing components. The constant portion entails instructions for utterance creation, while the changing part involves matching tokens with the corresponding entities.\\
\citep{hudevcek2023large} investigates the potential of exploiting instruction-tuned LLMs out-of-the-box in creating TODs. They also evaluate LLMs' performance in dialogue state tracking compared to supervised task-specific models. Their evaluations indicate that while LLMs may initially under-perform in dialogue state tracking, they effectively complete conversations if correct slots are provided. Their paper introduces a model for LLM-based TOD. In their presented model, TOD can be implemented in a zero-shot or few-shot manner. The dialogue state-tracking component comprises phases of prompting the LLM to extract the domain, and then extracting slots-related information from user input exploiting a created prompt. The extracted information then updates the conversation state, which is exploited to query the database. Then, a prompt is created to produce utterances, incorporating information from the belief state, retrieved entities, context, and user utterance.\\
In the few-shot manner, they introduce another phase at the initiation of the process. A context store is established by extracting correspondent conversations from the training collection centered on their congruence with the current conversation. Before prompt creation, items are retrieved from the context store; an item serves as a positive example, while another is corrupted and exploited as a negative example. They exploit certain instructed-tuned LLM models in zero-shot and few-shot manners to investigate the effectiveness of them in implementing their presented model. In addition, they exploit oracle belief states to evaluate the performance of their implemented TOD while excluding dialogue state tracking errors. They exploit MultiWOZ and schema-guided dataset in their experiments for evaluation purposes.\\
\citep{iizuka2023clarifying} mitigates the lack of conversation-level evaluation of dialogue systems constructed on LLMs. To complete this gap, they exploit MultiWOZ and conduct a human evaluation to measure the performance of dialogue systems constructed on the model presented in \citep{hudevcek2023large}. They are compared with other conversational systems including rule-based and TOATOD \citep{bang2023task}, an end-to-end TOD. Conducted evaluations illustrate that GPT4 performs comparably to rule-based systems and consistently outperforms other implemented conversational systems.\\
LLMs perform competently in creating utterances within chat environments, but their efficacy diminishes in complex conversations, such as medical or legal consultations. \citep{cao2023diaggpt} presents a model for proactive management of professional consultation conversations, leveraging the capabilities of LLMs. They claim this model works as a more clever and professional chatbots. The presented model comprises chat agent, topic manager, topic enricher, and context manager modules. Their presented model characterizes the topic as the primary concentration of each cycle in the conversation, and it characterizes task as the particular mission which is deemed complete upon confronting the whole predefined topics in the conversation.\\
In the topic evolution phase, the topic manager, working as primary module of their model, processes input encompassing the user query, chat context, action list, and the state of the topic stack. It then predicts the crucial action for topic evolution in the current cycle. In this model, each action corresponds to a particular program to implement an instruction. Predefined actions in their presented model include creating a topic, concluding the current topic, remaining within the current topic, and finalizing the current topic. Execution of the command corresponding to the predicted action enables topic transitions, including the maintenance of the topic stack which is a data structure to maintain conversation states. Then, the topic enricher, by processing the current topic and the output of the context manager, presents input for creating utterances which is the last phase and is enabled by the chat agent module.\\
\citep{bocklisch2024task} examines the intent-based NLU approach, which is the conventional model for creating industrial TODs comprising natural language understanding (NLU), dialogue management (DM), and natural language generation (NLG) components. In this approach, NLU is modeled as a classification task with predetermined intents as classes; consequently user utterances are categorized into their corresponding intents. While working with a collection of limited predefined intents can pose challenges in updating the intents. The essence of intents lies in their conversion to tasks, but not every user utterance can be precisely mapped to tasks or predefined intents.\\
\citep{bocklisch2024task} presents an LLM-based model for industrial TODs, which considers the potentials of in-context learning of LLMs, and mitigates the challenges of the intent-based NLU approach. The principal components of their presented model include business logic, dialogue understanding (DU), and conversation repair. In each conversation cycle, DU, informed by the entire conversation context, converts user utterances into a sequence of commands. The dialogue manager processes commands to update the dialogue stack, which prorates the conversation state. If conversation repair is warranted centered on the user utterance, the dialogue stack is updated centered on the corresponding improving pattern. \\\citep{Pittsburgh2023} presents a collaborative model that mitigates the paramount challenge of considering different user populations within taskbots. This model contains a state graph that prorates conversation and context to improve NLU. The presented model's DM is capable of creating clarifying questions to moderate misunderstandings in conversations. To create an inclusive multi-modal interface, they conduct collaborative efforts with many communities, including the deaf and hard of hearing (DHH). In addition, the NLG component integrates LLM-based and rule-based modules to diminish concerns of efficiency and privacy in creating utterances.\\
The paper \cite{xprotod} concentrates on the limitations of LLM-based, particularly in-context learning-based TODs. In those TODs, passive planning and non-iterative retrieval mechanism are challenging; consequently, the paper creates ProTOD to improve proactivity and task completion. ProTOD consists of an exploratory retrieval mechanism  and a passive-to-proactive policy planner. The retrieval mechanism is centered on chain-of-plan-execution to explore comprehensive corresponding information by exploiting collection of tools. If policy considered as the list of actions, these actions are not only influenced by context information but are also contingent on each other. In planning it indicates a collection of mandatory actions as the initial policy; consequently in the latter proactive phase, it integrates proactive actions like asking clarification questions. Integration considering the contingency between actions is multiphase process. in each phase, the model assesses if it is influential to integrate proactive actions, then creates an integrated policy and corresponding utterance.\\
The challenges in modular models of creating TODs is investigated in paper \citep{xetal}. In includes error propagation, customization costs, transferability, fault tolerance. Their paper presents a non-modular model which is centered on instruction template and LLM. To decrease customization costs and improve transferability, exploiting this instruction template in different issues is not complex. To improve adaptability in management of challenging conditions, there is not a lot of constraint in the instruction template; consequently, it does not limit the processing and changes in conversation. The instruction template initially indicates what the task category and TOD character are. Then  it contains the task information part, which explains the information like what the task is and which are the corresponding tools. The last part corresponds to the layout for passing information.
\subsection{Pro-activity and the Creation of Clarifying Questions }
\label{proactive}
Proactivity in a conversational system is the capability of managing a conversation by taking the initiative, predicting and considering the consequences of changes before implementing them, in completing purpose of the conversation. Proactive dialogue systems are capable of coordinating complex interactions and managing conversations toward a predetermined purpose \citep{deng2023survey}. While the LLMs enhance the capability of conversational agents to create utterances, LLMs are only trained to pursue instructions, and a lack of goal awareness poses challenges in current LLM-based conversational systems. \citep{deng2023rethinking} investigates LLM-based conversational agents and examines methods that have been presented to enhance goal awareness in proactive, non-collaborative, and multi-goal conversational systems. Proactivity is the most critical characteristic of goal awareness in intelligent conversational systems. In TODs, proactivity enables providing effective unprompted information. This capability is promising in cultivating enhanced collaboration between users and TODs toward the completion of tasks \citep{deng2023rethinking}.\\
\citep{deng2023prompting} inspects the capabilities of LLM-based dialogue systems in managing proactive conversations, including clarification, goal awareness, and non-collaboration conversations, as proactivity is paramount for eliciting clarifying questions, coordinating target-guided conversations, and managing non-collaborative exchanges. They present proactive chain-of-thought (ProCoT) prompting method to improve proactivity of LLM-based conversational systems through elevating their planning and taking initiative toward task completion. The ProCoT is conceptualized as p(t, a, r | D, C, A), D indicates the task background, C implies the context, A indicates the potential dialogue acts, and t explains the thought for the decision-making process concerning predicted action (e.g., ambiguity analysis of a user utterance in clarification conversations). In comparison with proactive prompting p(a, r | D, C, A), ProCoT involves a phase to check the effect of conducting predicted action on task completion. Their evaluations indicate that ProCoT improves the performance of clarification and target-guided conversations.\\
\citep{deng2022pacific} concentrates on proactive information-seeking conversations and introduces the proactive conversational question answering (PCQA) task, which combines conversational question answering (CQA) and clarification question generation (CQG). The PCQA tasks are decomposed into identifying ambiguity in the user query, proactively creating a clarifying question if the user query is identified as ambiguous, and answering the user query. Their paper introduces the PACIFIC dataset, crafted for PCQA over contexts of table and text in the finance domain. In addition, they present the UniPCQA multi-task learning model, which unifies whole subtasks of PCQA. While the performance of UniPCQA exploiting PACIFIC is not week, it is lower than that of experts, which indicates the essence of improving PCQA performance.\\
\citep{zamani2020generating} concentrates on creating clarifying questions within an information-seeking condition. They initially construct a clarification taxonomy by investigating Bing search logs which contribute to creating templates. Their paper presents certain models for creating clarification questions. \citep{sun2020adding} presents a method to integrate chitchat into TODs, which enables providing non-inquired information to enhance their interaction, interestingness, informativeness, and humanlikeness quality while maintaining task completion efficiency. It introduces a methodology for dataset creation and implements it to create a dataset with inserting chitchat for copious conversations of schema-guided dialogue (SGD) and MultiWOZ task-oriented datasets. In their paper, certain models are created. These models are trained on their created collection to produce chitchat-augmented task-oriented conversations. While this integrated chitchat information improved user experience, it probably contains limited effective information for the user. In contrast, \citep{chen2022ketod} provides a dataset to mitigate this issue, which considers integrating knowledge grounded chitchat into the conversation.
\subsection{ Investigation of Previous Works}
\label{xDemerits}
The performance of the last created LLMs leads to leveraging in-context learning of LLMs instead of calibrating pre-trained language models in creating many TODs. While the lack of a common baseline model for creating and comparing in-context learning-based TODs is challenging. \citep{hu2024dialight} concentrate on enabling the creation and evaluation of multi-lingual LLMs-based TODs. Their implementations are centered on TOD model comprising a dialogue state tracking, a database, and utterance creation components. Centered on this presumed model, their paper constructs instances of calibration-based and in-context learning-based TODs and evaluates their performance across multiple languages. In in-context learning-based TOD creation, it creates prompts for implementing the initial and last components. Centered on their evaluations , while the utterances created in the implemented in-context learning-based TOD are more diverse and likable, there is a gap between possible processing exploiting only in-context learning of LLMs and processing which is essential for creating TODs. \\
LLMs, trained on the corpus of internet, are capable of conducting utterance creation and language understanding tasks centered on precise instructions across different domains without imposing calibration. Challenges in task-oriented conversations encompass understanding, creating utterances, and planning. Planning in LLMs is challenging \citep{huang2024understanding}, while constituting a pivotal component of TODs.  The primary purpose of TODs is to complete a particular task—conventionally involving retrieving entities from a database centered on the conversational exchange between the user and the TOD \citep{polifroni2006learning}— which is challenging even with errorless language processing. While LLMs exploiting in-context learning commonly excel in utterance creation and language understanding, weak planning may lead to TODs creating utterances with commendable BLEU \citep{papineni2002bleu} scores that lack informativeness for TOD users.\\
In terms of planning, considering planning as producing a list of actions, the methods exploited in the LLM-based platforms to improve their planning can be categorized into task decomposition, plan selection, external module, reflection and memory. Task decomposition, centered on divide-and-conquer, decomposes the process into smaller parts, which are easier to plan and prevent infeasible and inefficient planning. The external module or external planer-aided planning method converts information that must be provided to the planning into intermediate representation leveraging an LLM. Then, an external planner is exploited to perform planning exploiting this information. The equation \ref{planner} corresponds to this process. With the environment as E and the task goal as g, $\Theta$ is the parameters of the LLM, P is the prompts, $\Phi$ indicates the external planner module, and h is the instituted information. This integration enhances LLM planning in terms of interpretability, conceptual completeness, and constancy. These methods are interconnected more than mutually exclusive \citep{huang2024understanding}.
\begin{align}
    h &= \text{formalize}(E, g; \Theta, P), \nonumber \\
    p &= \text{plan}(E, g, h; \Phi).
    \label{planner}
\end{align}
Proactive planning, which prevents error-prone conditions and conversation defeat, is possible through exploiting effective intermediate conversation information, but most of prior works in exploiting explicit intermediate information are limited to predefined slots, intents, query information and its products. Besides, in previous predefined slots and intent-based models, modeling preferences centered on extracting and exploiting the whole effective information from utterances is challenging \citep{bocklisch2024task,xpaperpre}.
Table \ref{xpaperslearning} illustrates the previous works investigated in this paper. It indicates if the prior work is concentrated on creating LLM-based model, this model is centered on in-context learning of LLMs or on pretraining and calibration of LLMs. It is possible to perceive if the investigated paper is concentrates on proactivity and clarification, if it contains entity-ranking mechanism, if it explicitly considers the intermediate information other than predefined slots, intents and query and its corresponding entity information in its planning, and if it is centered on the information state model in conversion management.

\begin{table}[h!]
\caption{ previous works,
column abbreviations exploited: 
\textbf{LLM-based} -- LLM-based models; 
\textbf{precalib} -- pretraining and calibrating; 
\textbf{ICL} -- in-context learning; 
\textbf{pro-clari} -- proactivity or clarification; 
\textbf{ent-rank} -- entity ranking; 
\textbf{expl-other} -- explicit other intermediate information 
(not only predefined slots, intents, query info); 
\textbf{info state } -- information state model.
}

\hspace*{-0.5cm}
\begin{tabular}[width=\textwidt]{ccccccc}
\toprule
\textbf{paper} & \multicolumn{2}{c}{\textbf{LLM-based}} & \textbf{ pro- }  & \textbf{ent- } & \textbf{  expl-   }  & \textbf{info   }  \\
\cmidrule{2-3}
   & \textbf{precalib  } & \textbf{ICL} &\textbf{ clari }  &\textbf{rank}  &\textbf{other }  &\textbf{state}\\
\midrule
    
Peng et al. \cite{peng2021soloist}& * &  & & & & \\

Hosseini-Asl et al. \cite{hosseini2020simple}& * &  & & & & \\

 Iizuka et al.\cite{iizuka2023clarifying}& * &  & && & \\

 Qin et al.\cite{qin2023modularized}& * &  & & & &  \\

 He et al. \cite{he2022galaxy}& * &  & & & & \\

 Kulhánek et al. \cite{kulhanek2021augpt}& * &  & & & & \\

 Su et al. \cite{su2022multi}& * &  & & & & \\

Yang, Li, and Quan\cite{yang2021ubar}& * &  & & & & \\

 Tabassum et al.\cite{Tech2023}& * &  & & & & \\

 Chung et al.\cite{chung-etal-2023-instructtods}&  & * & & & & \\

 Gao et al.\cite{gao-etal-2023-adaptive}&  & * & & & & \\

Hudeček, and Dusek\cite{hudevcek2023large}&  & * & & & & \\

Cao \cite{cao2023diaggpt}&  & * & & & & \\

Bocklisch et al.\cite{bocklisch2024task}&  & * & & & & \\

Xu et al.\cite{xetal}&  & * & & & & \\

Dong, Chen and Yang\cite{xprotod}\cite{xprotod}&  & * & *& & & \\

 Sicilia et al.\cite{Pittsburgh2023}&  & * & * & & &\\

 Deng et al.\cite{deng2023rethinking}&*  & * & *& & & \\

 Deng et al.\cite{deng2023survey} & * & *&   *& & &\\

 Deng et al. \cite{deng2022pacific}&  *&  &  *&  & &\\

 Deng et al.\cite{deng2023prompting}&  & * & *&  & &\\

  Sun et al.\cite{sun2020adding}&  *&  &* & & & \\

 Chen et al. \cite{chen2022ketod}&*  &  & *& & & \\

 Zamani et al.\cite{zamani2020generating}&  &  & *& & & \\

created model  &  & * & *& *& *& *\\

\botrule
\end{tabular}

\label{xpaperslearning}
\end{table}

\section{Method}
\label{Method}
In TOD, the primary purpose is to complete particular tasks centered on preferences. In these conversations, preferences illustrate the constraints of database queries. In the presented model in this paper, considering the limitations of TOD models centered on predefined slots and intents, preferences are modeled not only with predefined slots but also with a text\_part which contains the constraints beyond the predefined slots. This model enables to consider the whole preferences.\\
With these presumptions, it is possible to create conversational platform which   processes user inputs, retrieves database entities that are congruent to the predefined slots of the preferences, and indicates the order of these entities centered on their congruence to the text\_part of the preferences. The TOD then can conclude the conversation by presenting the ordered entities. While, deviations from this path are common, considering errors in preferences, limitations within the database entities, or entity rejection.\\

If the user inputs, "I am looking for a restaurant that serves canap in the east." and "canap" is incorrect or out-of-domain input for TOD, it can create a clarifying question, prompting the user to modify the incorrect or out-of-domain input. If the user input is, "can you help me find a restaurant that serves Italian food with a moderate price range please", and none of the user-indicated inputs are wrong or out-of-domain, but there is no entity in the database that is congruent with this combination of preferences. In this circumstance, instead of merely informing the user that there is no congruent entity , the TOD can provide effective information centered on the database entities and user preferences. It might inform the user of entities in the database with predefined slots such as: "price: cheap, food: Italian" or "price: moderate, food: European." In this way, it contributes to the modification of input preferences collaboratively.\\

With concentration on changes in informative elements across different conversation paths, many task-oriented conversations investigated. Consequently, centered on conducted investigation of conversations, it is possible to create limited informational components which are constructive in planning, and their changes portray the changes in conversation conditions, including but not limited to the predefined slots and text\_part in preferences. These include informational components that portray circumstances such as user input errors in predefined slots, and rejection of retrieved entities. Consequently, these limited informational components lead to limited possible Information states. Considering movement between possible information states, dialogue moves are created. The created informational components, and the dialogue moves are listed in Figure \ref{xinformationcomponents}.\\
Centered on information state approach to dialogue management, an update strategy is constructed. Creating an update strategy which is broken down, centered on the critical points of changing informational components enables explicitly exploiting identified informational components in planning. With the conventional calibration-based LLMs, it is impractical to calibrate each decomposed part of the model. While in-context learning of LLMs—inspired by task decomposition and external planer-aided planning methods —enables implementing this update strategy. The update strategy in the presented model is close to the position of an external planner ($\Phi$ ) in equation \ref{planner} .\\
\begin{figure}[H]
\begin{tcolorbox}[colback=white!5!white, colframe=black!75!black]

 Predefined\_slots\\
Text\_part\\
It\_Is\_required\_to\_update\_predefined\_slots\\
 It\_Is\_required\_to\_query\_database\\
 db\_query\_output\_list\_is\_updated\\
 query\_output\_list\_is\_empty\\
dialogue\_is\_completed\\
utterance\_to\_update\_predefined\_slot\\
 checked\_there\_is\_no\_other\_constraint\\
 user\_rejects\_output\\
 there\_is\_wrong\_or\_out\_of\_domain\_value\\
 wrong\_or\_out\_of\_domain\_values\_list\\
 user\_utterance\_index\\
 user\_other\_constraints\\
wrong\_ness\_within\_other\_constraints\_checked\\
user\_is\_informed\_there\_is\_no\_entity\_in\_db\\
 user\_is\_informed\_of\_db\_output\\
 domain\_caption\\
 extracted\_information\\
 \tcblower
update\_user\_preferences\\
check\_is\_there\_wrong\_or\_out\_of\_main\_value\\
create\_clarifying\_question\_InputIsWrong-or-Out-of-domain\\
create\_clarifying\_question\_QueryoutputIsEmpty\_OutputIsRejected\\
Inform\_user\_there\_is\_no\_entity\_in\_db\\
query\_database\\
 check\_the\_emptiness\_of\_query\_output\\
check\_if\_the\_user\_wants\_to\_enter\_more\_constraints\\
entity\_ranking\\
check\_if\_user\_rejects\_output\\
end\_dialogue\\
\end{tcolorbox}
 \caption{Informational Components, Dialogue Moves}
\label{xinformationcomponents}
\end{figure}
In \ref{xpresentedmodel} process and creation of utterances in the created model illustrated. In this model, the cycle corresponds to an interchange of utterances between user and conversation platform. There might be multiple phases for each cycle. In each phase, the update strategy, considering the information state of the conversation, indicates the later dialogue move which leads to performing procedures corresponding to that dialogue move. The product of these procedures passed to update strategy which, if the utterance is prepared, passes it to user and the current cycle ended. If not, it proceeds to the latter phase of current cycle.\\

\begin{tcolorbox}[colback=yellow!5!white,colframe=yellow!50!black,colbacktitle=yellow!75!black]
\begin{itemize}
\item user utterance ($U_x$): Input at conversation cycle x
\item information state ($ I_{x} $): TOD's current contextual understanding
\item dialogue move ($ D_{x} $): Possible TOD's actions/moves
\item update strategy ($U_x$): Identifies proper dialogue moves or presents created TOD utterance to user

\item procedures ($ P_{x_i} $): Procedures corresponding to particular dialogue move ($ D_{x_i} $)
\item  TOD utterance ($T_x$): Created utterance at cycle x of conversation
\end{itemize}
  \tcblower
 initially the information state is $ I_{0} $, \\
 at cycle x if conversation is not ended:
\begin{itemize}
 \item capture user input $U_x$
 \begin{itemize}
\item at i-th phase of cycle x:
\begin{itemize}
 \item update information state, continue processing with $ I_{x_i} $
\item perform update strategy $ M_{x_i} $( $ I_{x_i} $)
  \item identify latter dialogue move $ D_{x_i} $
 \item perform procedure $ P_{x_i} $( $ I_{x_i} $) corresponding to $ D_{x_i} $
  \item update information state
  \item if  $ T_{x} $ utterance is prepared, pass it to user
  \begin{itemize}
	\item if conversation is ended, proceed end phase, if not increment x, break the phases loop of cycle x, proceed with latter conversation cycle
     \end{itemize}
       \item if $ T_{x} $  is not prepared
        \begin{itemize}
         \item increment i, proceed to latter phase of cycle x
         \end{itemize}
\end{itemize}
\end{itemize}
\end{itemize}
    end
\label{xpresentedmodel}
\end{tcolorbox}	

In this model, each dialogue move is mapped to a list of procedures; the title of dialogue move is indicated centered on the title of the main procedure in its correspondent procedure list. The processing in dialogue moves' correspondent procedures include language processing, database retrieval, and informational component configurations. This part provides the explanation of the main procedure in the list of procedures corresponding to each dialogue move.\\
\textbf{update\_user\_preferences} 
Exploiting domain, predefined slots characterizations and predefined slots possible configurations for that domain, extract user configuration for each predefined slot.
Certify these configurations against the domain of the predefined slots.
If a configuration is congruent to the domain, impart it to the corresponding predefined\_slot component.
If the user enters a predefined slot's configuration, but it is incorrect or out\_of\_domain, create another component for that predefined slot ending with wrong\_or\_out\_of\_domain, and impart that extracted configuration into it. Insert constraints in the user input that are not imparted to predefined slots component to the text\_part component.\\

\begin{figure}[h!]

  \centering
\includegraphics[width=\textwidth]{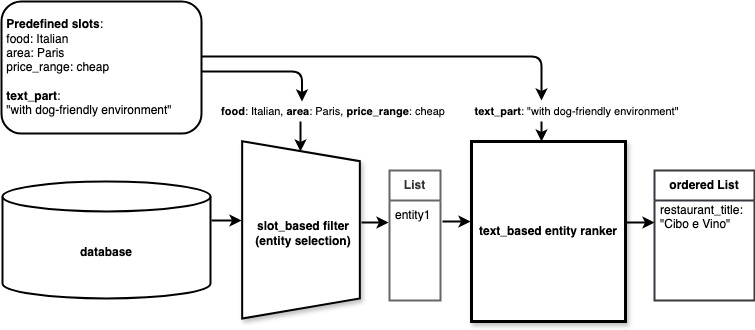}
 \caption{ Entity Search Model}
  \label{components}
\end{figure}

\textbf{check\_is\_there\_wrong\_or\_out\_of\_main\_value}
Process the output of update\_user\_preferences; if the configurations corresponding to predefined slot components, which their caption ending with \\wrong\_or\_out\_of\_domain are not no content, insert their configuration to the wrong\_or\_out\_of\_domain\_values\_list informational component. \\Consequently, if there is any component with wrong\_or\_out\_of\_domain ending, which is not no content,\\ set the there\_is\_wrong\_or\_out\_of\_domain\_value to true; else, it is false.\\
\textbf{ create\_clarifying\_question\_InputIsWrong-or-Out-of-domain}
Create a clarification question, which informs the user about the entered configurations in the wrong\_or\_out\_of\_domain\_values\_list, identified as incorrect or out-of-domain and must be changed if not they will be considered as 'none'. Put this question in utterance\_to\_update\_predefined\_slot informational component.\\
\textbf{query\_database} 
Retrieve the entities in the database that are congruent to the extracted predefined slot configurations and \\put them in the db\_query\_output\_list\_is\_updated informational component.\\
 \textbf{check\_the\_emptiness\_of\_query\_output}
Investigate the informational component db\_query\_output\_list\_is\_updated.\\ If there are no entities in it, set query\_output\_list\_is\_empty to true; otherwise, set it to false.\\
 \textbf{Inform\_user\_there\_is\_no\_entity\_in\_db}
Inform the user, what their preferences, inferred by TOD, are that are currently extracted and are in predefined slots and text\_part informational components, and inform the user that, there are no entities in the database that are congruent to these preferences.\\ Put true in user\_is\_informed\_there\_is\_no\_entity\_in\_db.\\
\textbf{create\_clarifying\_question\_QueryoutputIsEmpty\_OutputIsRejected}\\
Create a clarifying question in conditions that the database query output is empty, or the user rejects the provided database entities; not only inquiry the user to change configurations of preferences, but also proactively present information which enables the user to change configurations. Create a list (user\_item\_list) of potential configurations centered on existing predefined slot configurations and database entities.\\
\begin{itemize}
      \item Concerning each predefined slot with a configuration that is not 'none', identify the most comparable configurations in the database (preserving the other predefined slots unchanged).
    \item If the output of the query created exploiting this configuration is not empty, insert the configuration of predefined slots as an item into the user\_item\_list.
\end{itemize}   
After processing whole not 'none' predefined\_slots, create a question which presents the configurations in the user\_item\_list to the user and prompts the user to change the predefined slots  centered on the information provided.\\ Put the created question in the utterance\_to\_update\_predefined\_slot informational component.\\
\textbf{check\_if\_the\_user\_wants\_to\_enter\_more\_constraints}
Inform the user what the predefined\_slots and text\_part informational components maintain as the current user preference. Inquire if the user wants to update constraints. Process input; If it contains other constraints, put the input in the user\_other\_constraints informational component and set 'output' to true. Otherwise, set 'output' to false.\\
\textbf{entity\_ranking} 
Prioritize the entities in the db\_query\_output\_list\_is\_updated centered on the text\_part informational component. \\Update db\_query\_output\_list\_is\_updated. To prioritize database entities centered on constrains in text\_part informational component, an entity ranking mechanism created in \citep{saedi2024entity} has been exploited.\\
The exploited entity ranking mechanism is a component of an entity search that models user preferences through predefined slots and text\_part. In this entity search, entities are initially filtered centered on predefined slots. The order is then indicated by comparing the text\_part of user preferences (in that work text\_part was the concatenation of user\_review captions for each entity) exploiting a created transformer-based entity\_ranker.\\
The procedures corresponding to this dialogue move, do not include the initial filtering phase, as an equivalent mechanism is earlier exploited in the procedure corresponding to query\_database dialogue move. Considering that, in the update strategy, performing procedures corresponding to query\_database dialogue move is a prerequisite for the entity\_ranking dialogue move. Instead, entity\_ranking concentrates on indicating the order of the filtered entities within the db\_query\_output\_list\_is\_updated, exploiting the text-based entity\_ranker component of entity search illustrated in Figure \ref{components}.\\
Eventually, db\_query\_output\_list\_is\_updated is updated with the ordered list, the user is informed about the content of the db\_query\_output\_list\_is\_updated and whether want to reject these items or not, user clarification is issued as the output of these procedures, and user\_is\_informed\_of\_db\_output is set to true.\\

 \textbf{check\_if\_user\_rejects\_output}
Process the user utterance (which is the output of entity\_ranking) to illustrate if the user rejects the presented list of entities.\\ Set user\_rejects\_output to true if the user rejects the list; otherwise, set it to false.\\

\textbf{end\_dialogue}
Conclude the conversation by informing the user that the conversation has ended. Set dialogue\_is\_completed to true.\\

\begin{figure}[h!]
 
    \includegraphics[width=\textwidth]{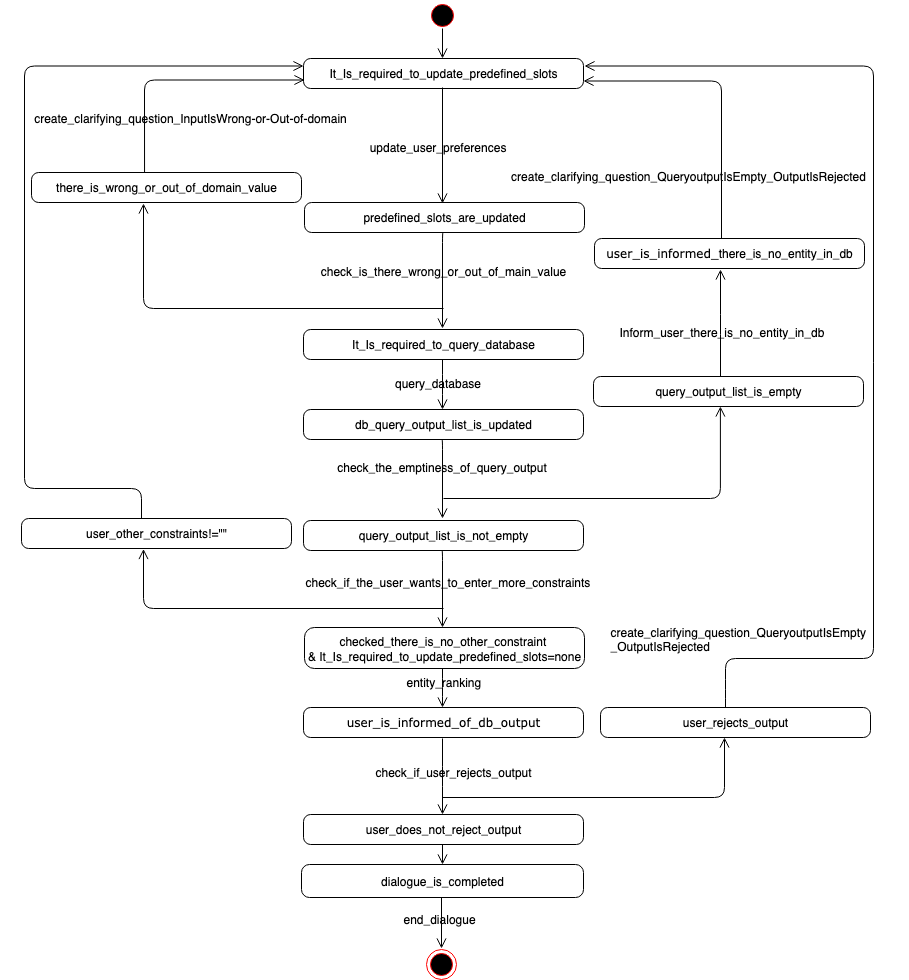}
    \caption{Update Strategy: Dialogue Moves \& Changes in Information States in Conversation}
    \label{ximageupdatestrategy}
\end{figure}

Figure \ref{ximageupdatestrategy} illustrates the update strategy with a concentration on dialogue moves and changes in information state. 

The pseudocode of update strategy, which indicates its implementation to manage conversations is presented in \ref{xupdatestrategy}. The explanations of managing conversations centered on this update strategy are provided. \\
\begin{algorithm}
\xtiny

\caption{Implementation of Update Strategy to Manage Conversations}
\label{xupdatestrategy}
\begin{algorithmic} 
\While{not dialogue\_is\_completed}
    \If{It\_Is\_required\_to\_update\_predefined\_slots}
        \State inp $\gets$ ""

        \If{user\_other\_constraints $\neq$ "" and not wrong\_ness\_within\_other\_constraints\_checked}
            \State inp $\gets$ user\_other\_constraints
            \State \textbf{Print} "TOD: " + utterance\_to\_update\_predefined\_slot
        \Else
            \State inp $\gets$ user\_input(utterance\_to\_update\_predefined\_slot)
        \EndIf

        \State user\_is\_informed\_there\_is\_no\_entity\_in\_db $\gets$ None
        \State update\_out $\gets$ update\_user\_preferences(inp, output\_list\_domain)
        \State out\_check $\gets$ check\_is\_there\_wrong\_or\_out\_of\_main\_value(update\_out)
        \State there\_is\_wrong\_or\_out\_of\_domain\_value $\gets$ out\_check
        \State It\_Is\_required\_to\_query\_database $\gets$ not out\_check
        \State It\_Is\_required\_to\_update\_predefined\_slots $\gets$ None

        \If{there\_is\_wrong\_or\_out\_of\_domain\_value}
            \State create\_clarifying\_question\_InputIsWrong-or-Out-of-domain()
            \State It\_Is\_required\_to\_update\_predefined\_slots $\gets$ True
            \If{user\_other\_constraints $\neq$ ""}
                \State wrong\_ness\_within\_other\_constraints\_checked $\gets$ True
            \EndIf
            \State there\_is\_wrong\_or\_out\_of\_domain\_value $\gets$ None
            \State wrong\_or\_out\_of\_domain\_values\_list $\gets$ []
        \Else
            \State It\_Is\_required\_to\_update\_predefined\_slots $\gets$ None
            \State It\_Is\_required\_to\_query\_database $\gets$ True
        \EndIf
        \State there\_is\_wrong\_or\_out\_of\_domain\_value $\gets$ None
        \State wrong\_or\_out\_of\_domain\_values\_list $\gets$ []
    \EndIf

    \If{It\_Is\_required\_to\_query\_database}
        \State query\_database(only\_check=False)
        \State check\_the\_emptiness\_of\_query\_output()
        \If{query\_output\_list\_is\_empty}
            \State Inform\_user\_there\_is\_no\_entity\_in\_db()
            \State create\_clarifying\_question\_QueryoutputIsEmpty\_OutputIsRejected(output\_list\_domain)
            \State It\_Is\_required\_to\_update\_predefined\_slots $\gets$ True
        \Else
            \State out\_check $\gets$ check\_if\_the\_user\_wants\_to\_enter\_more\_constraints()
            \State checked\_there\_is\_no\_other\_constraint $\gets$ not out\_check
            \If{not checked\_there\_is\_no\_other\_constraint}
                \State It\_Is\_required\_to\_update\_predefined\_slots $\gets$ True
                \State checked\_there\_is\_no\_other\_constraint $\gets$ True
            \EndIf
        \EndIf
        \State query\_output\_list\_is\_empty $\gets$ None
        \State It\_Is\_required\_to\_query\_database $\gets$ None
    \EndIf

    \If{It\_Is\_required\_to\_update\_predefined\_slots == None and checked\_there\_is\_no\_other\_constraint}
        \State user\_other\_constraints $\gets$ ""
        \State user\_inp $\gets$ entity\_ranking()
        \State check\_if\_user\_rejects\_output(user\_inp)
        \If{user\_rejects\_output}
            \State create\_clarifying\_question\_QueryoutputIsEmpty\_OutputIsRejected()
            \State It\_Is\_required\_to\_update\_predefined\_slots $\gets$ True
        \Else
            \State dialogue\_is\_completed $\gets$ True
        \EndIf
        \State user\_rejects\_output $\gets$ None
        \State checked\_there\_is\_no\_other\_constraint $\gets$ None
        \State User\_other\_constraint $\gets$ ""
    \EndIf
    \If{dialogue\_is\_completed}
        \State End\_dialogue()
        \State \textbf{Print} Predefined\_slots
    \EndIf
\EndWhile
\end{algorithmic}
\end{algorithm}

Initially, in the information state, It\_Is\_required\_to\_update\_predefined\_slots is true.

\textbf{ If It\_Is\_required\_to\_update\_predefined\_slots is true,}
perform procedures corresponding to update\_user\_preferences\\  which updates the configuration of Predefined slot and Text\_part. Then perform procedures corresponding to dialogue move of \\check\_is\_there\_wrong\_or\_out\_of\_main\_value to update informational components there\_is\_wrong\_or\_out\_of\_domain\_value and \\ wrong\_or\_out\_of\_domain\_values\_list of the information state.\\
\begin{itemize}
 \item  If there\_is\_wrong\_or\_out\_of\_domain\_value is false,\\
 set the It\_Is\_required\_to\_query\_database to true.\\Put 'none' in It\_Is\_required\_to\_update\_predefined\_slots.
   \item Otherwise, if there\_is\_wrong\_or\_out\_of\_domain\_value is true, \\
   perform procedures corresponding to \\create\_clarifying\_question\_InputIsWrong-or-Out-of-domain, which creates a clarifying question exploiting the information in \\wrong\_or\_out\_of\_domain\_values\_list and puts the clarifying question in utterance\_to\_update\_predefined\_slot.\\ Then set It\_Is\_required\_to\_update\_predefined\_slots to true. 
  \end{itemize}
 Put 'none' in there\_is\_wrong\_or\_out\_of\_domain\_value, and [] in \\wrong\_or\_out\_of\_domain\_values\_list.\\
 \textbf{If It\_Is\_required\_to\_query\_database is true,}
 perform procedures corresponding to the query\_database which preserves the output of query in\\ db\_query\_output\_list\_is\_updated. Then perform procedures corresponding to check\_the\_emptiness\_of\_query\_output\\ to update query\_output\_list\_is\_empty.
 \begin{itemize}
 \item If query\_output\_list\_is\_empty is true,\\
 perform procedures corresponding to Inform\_user\_there\_is\_no\_entity\_in\_db and create\_clarifying\_question\_QueryoutputIsEmpty\_OutputIsRejected, which informs user there is no congruent entity in the database, \\create a clarifying question, and put created question in \\utterance\_to\_update\_predefined\_slot. \\Then, set It\_Is\_required\_to\_update\_predefined\_slots to true. 
    \item If query\_output\_list\_is\_empty is false, perform procedures corresponding to check\_if\_the\_user\_wants\_to\_enter\_more\_constraints,\\ which informs the user of the current extracted preferences and whether the user wants to update them or not. 
    Centered on user clarification, if there are no other constraints, 'output' is false; but if there are other constraints, 'output' is true, then \\put not 'output' in checked\_there\_is\_no\_other\_constraint.
     \begin{itemize}
     \item If checked\_there\_is\_no\_other\_constraint is false,\\
        set It\_Is\_required\_to\_update\_predefined\_slots and \\checked\_there\_is\_no\_other\_constraint to true; 
   \end{itemize}
   \end{itemize}
 Put 'none' in It\_Is\_required\_to\_query\_database and query\_output\_list\_is\_empty.\\

  \textbf{If It\_Is\_required\_to\_update\_predefined\_slots is 'none', and\\checked\_there\_is\_no\_other\_constraint is true,}
  perform procedures corresponding to entity\_ranking, which indicate the order of entities in\\ db\_query\_output\_list\_is\_updated centered on the Text\_part of user preferences, update db\_query\_output\_list\_is\_updated, inform user what these entities are, and whether user rejects them or not. Put user clarification in the 'output'. Then set user\_is\_informed\_of\_db\_output to true.
  Perform procedures corresponding to check\_if\_user\_rejects\_output, which centered on the 'output' of entity\_ranking, updates user\_rejects\_output.\\
 \begin{itemize}
  \item If user\_rejects\_output is true, perform procedures\\ corresponding to \\create\_clarifying\_question\_QueryoutputIsEmpty\_OutputIsRejected, which creates a clarifying question, and \\puts it in utterance\_to\_update\_predefined\_slot.\\ Then, set It\_Is\_required\_to\_update\_predefined\_slots to true. 
  \item user\_rejects\_output is false, set dialogue\_is\_completed to true.
  \end{itemize}

  Put 'none' in checked\_there\_is\_no\_other\_constraint and user\_rejects\_output, and put "" in user\_other\_constraints.\\
  
 \textbf{If dialogue\_is\_completed is true,} perform procedures corresponding to End\_dialogue.\\
\section{Experiments}
\label{Experiments}
In this part, the information concerning the implementation of the presented model and the experiments carried out is presented in \ref{x_Experiments}, including evaluation metrics in \ref{eval_metrics} and the dataset in \ref{x_testdata}.  Consequently, the explanation of the performance and comparison with previous works is provided in \ref{Performance Examination}.
\subsection{Comprehensive Breakdown of Implementation and Experiments  }
\label{x_Experiments}
\paragraph{\textbf{Evaluation Metrics}}
\label{eval_metrics}
The evaluation of conversational systems can concentrate on their external behavior, internal capability, or impact on users and the processes they are a part of. The elements being evaluated are called constructs, which are operationalized to identify measurement methods. The construct can be operationalized in multiple ways. \citep{braggaar2023evaluating} investigates the constructs and their operationalizations for evaluating conversational systems, particularly in customer service context. It categorizes the constructs into intrinsic evaluation, and system-in-context classes. \\
The system-in-context construct is mainly created for TODs. It includes task success/effectiveness, usability, and user experience classes. Compared to intrinsic evaluation, most system-in-context constructs are examined through human evaluation, except for task completion, for which some mechanized evaluation methods exist \citep {braggaar2023evaluating}. \\
The primary purpose of a TOD is task completion, which can be measured objectively centered on the inform and success metrics. While human evaluation can be effective, it is commonly time-consuming and complex considering potential biases and differences in opinions. Therefore, in this paper, which concentrates primarily on task completion, TODs are evaluated and compared centered on objective metrics of task completion.\\
Inform and success, which correspond to the task completion (task success) construct of the system-in-context category, are the most common measures in previous TOD works. Inform measures the percentage of conversations in which the entities provided by the TOD do not contradict the current information state (dialogue state) and user goals. The success rate measures the percentage of conversations in which the TOD provides the whole user inquired information. In this paper, which concentrates on improving task completion, the whole congruent entities are presented to the user, including their entire information in the database in a table; therefore, evaluating inform will be equivalent to evaluating success \citep{yang2021ubar,peng2021soloist,gao-etal-2023-adaptive,he2022galaxy,chung-etal-2023-instructtods,kulhanek2021augpt}.
\paragraph{\textbf{Dataset}}
\label{x_testdata}
The MultiWOZ dataset contains task-oriented conversations, contributing to the training and evaluation of TODs. To evaluate the implemented system, we exploited the MultiWOZ 2.2 \citep{zang2020multiwoz} dataset, which is divided into training, development, and test subsets. In this paper, the test part of this dataset is exploited to investigate the task completion capabilities of the implemented conversational system. While, since the test part contains multi-domain conversations which are outside the extent of the paper's problem, conversations that cover more than a domain at a time are filtered out. The exploited collection of conversations includes conversations in the domains of hotel, train, attraction, and restaurants. In the exploited MultiWOZ dataset, there is a schema that indicates the predefined slots, their characterizations, and possible configurations for some of the predefined slots of each domain. Table \ref{extracted_info} illustrates this information for the restaurant domain.\\

\begin{table}[]

\caption{ Instance of MultiWOZ Metadata }
  
\small

    \begin{tabular}{p{1.5cm}p{5cm}p{5cm}}   
         \toprule
    \textbf{caption} & \textbf{characterization} & \textbf{possible configurations}  \\
  \midrule
    pricerange & Price budget for the restaurant & cheap, expensive, moderate  \\

    area & Area or place of the restaurant & centre, east, north, south, west  \\

    food & The cuisine of the restaurant you are looking for & not\_limited \\

    name & Name of the restaurant & not\_limited \\

    bookday & Day of the restaurant booking & monday, tuesday, wednesday, thursday, friday, saturday, sunday  \\

    bookpeople & How many people for the restaurant reservation & 1, 2, 3, 4, 5, 6, 7, 8  \\

    booktime & Time of the restaurant booking & extracted from corresponding column in table of database \\

    address & Address of the restaurant & extracted from corresponding column in table of database  \\

    phone & Phone number of the restaurant & extracted from corresponding column in table of database  \\

    postcode & Postal code of the restaurant & extracted from corresponding column in table of database \\

    ref & Reference number of the restaurant booking &extracted from corresponding column in table of database  \\
 \botrule
    \end{tabular}%
    
    \label{extracted_info}
\end{table}
In the exploited MultiWOZ dataset, there is a database table for each domain that contains information about entities. If the possible configurations for predefined slots are not indicated in the metadata, we extract them from the domain's corresponding database. The captions of predefined slots and the corresponding columns in the database might differ; therefore, we mitigate these inconsistencies in the pre-processing phase. Eventually, we compile the information for each domain into a dictionary, which includes predefined slots, their characterization and possible configurations, and the information of database entities. The information in this dictionary corresponds to the informational components of the unchanged part of the information state and is exploited by the implemented TOD, centered on the domain in the conversation.\\

The implemented method is domain-independent but does not conduct domain identification. In task completion models on entity search platforms, users commonly indicate the domain initially and then peruse information within that indicated domain. In this paper, the domain-dependent information—such as predefined slots characterizations, and corresponding databases— is configured in pre-processing. We will then perform the constant program with certain domain-dependent parameters configured in pre-processing. The LLM exploited in this paper's implementations is GPT-4o.\\
To clarify the implementation of created procedure, the pseudocode of update\_user\_preferences procedure is illustrated in \ref{xupdate_user_preferences}. It indicates that, provided domain and characterization of predefined slots inserted in their corresponding element in the prompt template, a domain independent prompt is consequently created to process user utterances and extract user preferences, including predefined slots and text\_part, This process also identifies incorrect or out-of-domain configurations and maintains them to configure their corresponding informational components later in the corespondent procedures of check\_is\_there\_wrong\_or\_out\_of\_main\_value dialogue move. The prompt includes instructions which indicate the output illustration. To ensure clarity, an example indicating the whole process is inserted into the prompt. Then, informational components corresponding to the predefined slots and text\_part in information state will be updated centered on the extracted information.\\

\begin{algorithm}

\xtiny
\caption{Update User Preferences}
\begin{algorithmic}
\Procedure{update\_user\_preferences}{input}

    \State domain $\gets$ information\_state.domain\_caption
    \State extracted\_info $\gets$ 
    information\_state.extracted\_information[domain]
    \State predefined\_slots\_caption\_list $\gets$ extracted\_info [captions]
    \State predefined\_slots\_configs\_list $\gets$ extracted\_info [configurations]
    \State predefined\_slots\_chars\_list $\gets$extracted\_info[characterizations]
    \State prompt $\gets$ "Understanding User Input and Slot Extraction for user Queries in \{domain\} domain"
    \For{each $i \gets 0$ to len(predefined\_slots\_caption\_list) - 1}
        \State pre\_char $\gets$ predefined\_slots\_chars\_list[i]
        \State predefined\_slot $\gets$ predefined\_slots\_caption\_list[i]
        \State pre\_config$\gets$ predefined\_slots\_configs\_list[i]
        \If {predefined\_slot $\neq$ "hotel-type"}
           \State prompt $\gets$ prompt + "Instructions for Extraction of \{predefined\_slot\} which indicate \{pre\_char\}:"
            \State prompt $\gets$ prompt + "- Provided a list of \{predefined\_slot\}-domain values: \{pre\_config\}"
            \State prompt $\gets$ prompt + "- Extract \{predefined\_slot\} values from the query \{input\} excluding any prefixes or suffixes like \{predefined\_slot\}."
            \State prompt $\gets$ prompt + "- If the \{pre\_config\} is empty, assign extracted \{predefined\_slot\} value to the \{predefined\_slot\}-variable and \{predefined\_slot\}-wrong-or-out-of-domain is 'None'"
            \State  prompt $\gets$ prompt + "- otherwise if \{pre\_config\}  is not empty, If and only if  I emphasis that if and only if at least one extracted \{predefined\_slot\} value exist with the same typo (ignoring case differences and without
                   \{predefined\_slot\} prefix/suffix,in addition for numbers consider their writing in letters and numbers the same) in the \{predefined\_slot\}-domain list which is  \{pre\_config\}, assign it to the
                  \{predefined\_slot\}-variable and \{predefined\_slot\}-wrong-or-out-of-domain is 'None'(not null)"
            \State   prompt $\gets$ prompt + "- Otherwise, if  \{pre\_config\} is not empty and there are extracted \{predefined\_slot\}  values in  \{input\}  but non of them exist with the same typo(ignoring case differences and without \{predefined\_slot\} prefix/suffix,in addition for numbers consider their writing in letters and numbers the same) in the \{predefined\_slot\}-domain list which is  \{pre\_config\},the value of \{predefined\_slot\}-variable is 'None'
                put the extracted value for \{predefined\_slot\} which is wrong or out-of-domain value, in {predefined\_slot}-wrong-or-out-of-domain"        
        \EndIf
    \EndFor
    \State prompt $\gets$ prompt + " put the information of variables in a Python dictionary and only display the dictionary (without any surrounding characters like triple quotes (''') or code block delimiters.) as output. the output dictionary consists of information of one variable in each line, information of each variable includes: "
    \State prompt\_x $\gets$ "Extract all attributes of \{domain\} from \{input\} and put them in text\_part\_variable, then remove the stop-words and the "
     \For{each $i \gets 0$ to len(predefined\_slots\_caption\_list) - 1 }
        \State predefined\_slot $\gets$ predefined\_slots\_caption\_list[i]
        \State prompt $\gets$ prompt + "\{predefined\_slot\} and the value of \{predefined\_slot\}-variable extracted from \{input\} (without any change for example do not change "None" value to other forms like null), \{predefined\_slot\}-wrong-or-out-of-domain and its value,"
        \State prompt\_x $\gets$ prompt\_x + "\{predefined\_slot\}-variable, and prefix and suffixes such as \{predefined\_slot\}, "
     \EndFor
    \State prompt\_x $\gets$ prompt\_x + "from these text\_part-variable, put one extra line in the output for text\_part which consist of "text\_part:text\_part\_value""
    \State prompt $\gets$ prompt + prompt\_x
    \State prompt $\gets$ prompt + "For example, if the domain is \{-\}, and the input is \{-\}',... the output would be:\{-\}"
    \State messages $\gets$ [{"role": "system", "content": prompt}]
    \State response $\gets$ client.chat.completions.create(model=llm\_model, messages=messages)
    \State generated\_list $\gets$ response.choices[0].message.content
    \State output $\gets$ eval(generated\_list)
    \For{(x, y) in output.items()}
        \If {type(y) is list} \State y $\gets$ y[0]
        \EndIf
        \For{(z, w) in self.information\_state.Predefined\_slots.items()}
            \If{x == z and y != 'None' and y != None} \State self.information\_state.Predefined\_slots[x] $\gets$ y
            \EndIf
        \EndFor
        \If {x == "text\_part"} \State self.information\_state.Text\_part $\gets$ y + " " + self.information\_state.Text\_part
        \EndIf
    \EndFor
\EndProcedure

\end{algorithmic}
\label{xupdate_user_preferences}
\end{algorithm}
The conversations in MultiWOZ consist of the domain caption, the utterances of the user and system, and the conversation state (including predefined slots configurations) in each cycle. In the evaluation of TODs exploiting a dataset, user utterances are constant and do not change centered on the current conversation with the implemented TODs, which presents a challenge. Consequently, a portion of conversations are completed before the whole user utterances in MultiWOZ are consumed; the example in Figure \ref{xunusedturns} illustrates this. Considering this condition and the characterization of "inform,"  the indicated MultiWOZ conversation state from the last exploited conversation cycle (whose user utterance is consumed in the conversation with the implemented TOD) is considered as the labeled user goal. Then, extracted predefined slots are compared with it.\\
\begin{figure}[H]

\small
\begin{tabularx}{\textwidth}{|p{0.6cm}|X|}
\hline
\rotatebox{-90}{\tiny{MultiWOZ Utterances} }
 &
\parbox[t]{\hsize}{
\begin{itemize}

 \item 'I am looking for a place to stay. The hotel should have a star of 2 and should  \item' be in the moderate price range'
 \item 'I don't have a preference, actually. Which one do you recommend?'
 \item 'I choose the ashley hotel. What is their address, please?'
 \item 'No, I just need the address.', 'Ok thank you that is all I needed today.'
\end{itemize}
} \\
\hline
TOD & enter query \\
\hline
User & I am looking for a place to stay. The hotel should have a star of 2 and should be in the moderate price range. \\
\hline
TOD & The values are: Price Range: moderate, Stars: 2. Are there any other constraints besides the ones already mentioned? \\
\hline
User & I don't have a preference, actually. Which one do you recommend? \\
\hline
TOD & Certainly! Here is the information presented in a table format:\newline
\parbox[t]{\hsize}{
\begin{tabularx}{\hsize}{c|c|c|c|c|c|c|c|c|c|c|c|c|c}
\hline
\tiny{\rotatebox{-90}{No.}} & \tiny{\rotatebox{-90}{Address}} & \tiny{\rotatebox{-90}{Location}} & \tiny{\rotatebox{-90}{Parking}} & \tiny{\rotatebox{-90}{Wifi}} & \tiny{\rotatebox{-90}{Checkin}} & \tiny{\rotatebox{-90}{Checkout}} & \tiny{Name} & \tiny{\rotatebox{-90}{Phone}} & \tiny{\rotatebox{-90}{PC}} & \tiny{Price}& \tiny{\rotatebox{-90}{Breakfast}} & \tiny{Rating} & \tiny{Type} \\
\hline
\tiny{x} & \tiny{x} & \tiny{x} & \tiny{x} & \tiny{x} & \tiny{x} & \tiny{x} & \tiny{Lovell Lodge} & \tiny{x} & \tiny{x} & \tiny{Mod }& \tiny{x} & \tiny{2} & \tiny{Hotel} \\
\hline
\tiny{x} & \tiny{x} & \tiny{x} & \tiny{x} & \tiny{x} & \tiny{x} & \tiny{x} & \tiny{Ashley Hotel} & \tiny{x} & \tiny{x} & \tiny{Mod} & \tiny{x} & \tiny{2} & \tiny{Hotel} \\
\hline
\end{tabularx}} 
Please review the table and indicate if you would like to reject the retrieved items or not.\\
\hline
User & I choose the Ashley Hotel. What is their address, please? \\
\hline
TOD &  I'm sorry, but this is the end of our dialogue. If you have any more questions or need further assistance in the future, feel free to ask. Have a great day! \\
\hline

\multicolumn{2}{|X|}{%
\parbox[t]{\hsize}{
Extracted\_info:\{ 'pricerange': 'moderate', 'stars': '2'\} \newline
MultiWOZ\_info: \{'name': ['ashley hotel'], 'pricerange': ['moderate'], 'stars': ['2'], 'type': ['hotel']\}
}
} \\
\hline

\end{tabularx}
\caption{Conversation Containing Excluded Utterances}
\label{xunusedturns}
\end{figure}
Within MultiWOZ in a domain, the domain caption and a configuration of a predefined slot are equal, which leads to confusion in two conversations. The conversation illustrated in Figure \ref{xconfusion} is an example of this condition. In this conversation, the user indicates the predefined slot type is hotel and therefore it is not a guesthouse. While, there is no entity in the database matching the user’s preferences. In contrast to user's clarification, the MultiWOZ conversation presents a guesthouse instead of a hotel because it does not consider the predefined slot for type. By ignoring the predefined slot of type in this domain in the predefined slots extraction, it works correctly. Therefore, in the procedures corresponding to update\_user\_preferences dialogue move, which are illustrated in \label{update_user_preferences}, the instruction is modified to ignore the predefined\_slot of type in the user utterances in this domain to mitigate this dataset issue.\\
\begin{figure}[H]

\small

\begin{tabularx}{\textwidth}{|p{0.6cm}|X|}
\hline
TOD & enter query \\
\hline
User & I am looking for a place to stay. The \textbf{hotel} should be in the east and should include free wifi. \\
\hline
TOD & \parbox[t]{\hsize}{The values provided are: Area: East, Internet: Free. You have mentioned that you need a place to stay in the east area and it should include free wifi. Are there any other constraints besides the ones already mentioned?
} \\
\hline
User & The rating should be 4 stars and I want free parking to be included. \\
\hline
TOD & \parbox[t]{\hsize}{I am updating your preferences based on the entered constraints. Please wait a moment while I process the changes. The values are:
Area: East, Internet: Free, Parking: Free, Stars: 4. Is there any other constraints besides the ones already mentioned?
} \\
\hline
User & No, will you just book me something for Wednesday for 1 person for 5 nights? \\
\hline
TOD & \parbox[t]{\hsize}{I'm updating your user preferences based on the constraints you've provided. This process might take a few moments. If you have any additional specifications or changes you'd like to make, please feel free to let me know. The values are: Area: East, Internet: Free, Parking: Free, Book Day: Wednesday, Book People: 1, Book Stay: 5 nights, Stars: 4 . I will book a place to stay on Wednesday for 1 person for 5 nights, with a 4-star rating, and it should include free WiFi.
Are there any other constraints besides the ones already mentioned?
} \\
\hline
User & That's all I need, thank you! \\
\hline
TOD & Certainly! Here is the information presented in a table format: 
\newline
\parbox[t]{\hsize}{
\begin{tabularx}{\hsize}{c|c|c|c|c|c|c|c|c|c|c|c|c|c|c}
\hline
\rotatebox{-90}{\tiny{ID}} & \rotatebox{-90}{\tiny{Address}} & \rotatebox{-90}{\tiny{Area}} & \rotatebox{-90}{\tiny{Parking}} & \rotatebox{-90}{\tiny{Internet}} & \rotatebox{-90}{\tiny{MinPrice}} & \rotatebox{-90}{\tiny{MaxPrice}} & \rotatebox{-90}{\tiny{AvgPrice}} & \rotatebox{-90}{\tiny{Name}} & \rotatebox{-90}{\tiny{Phone}} & \rotatebox{-90}{\tiny{Postcode}} & \rotatebox{-90}{\tiny{Price}} & \rotatebox{-90}{\tiny{ChildPerm}} & \rotatebox{-90}{\tiny{Rating}} & \rotatebox{-90}{\tiny{Type}} \\
\hline
\tiny{x} & \tiny{x} & \tiny{east} & \tiny{exist} & \tiny{exist} & \tiny{x} & \tiny{x} & \tiny{x} & \tiny{x} & \tiny{x} & \tiny{x} & \tiny{mod} & \tiny{x} & \tiny{4} & \tiny{guest} \\
\hline
\tiny{x} & \tiny{x} & \tiny{east} & \tiny{exist} & \tiny{exist} & \tiny{x} & \tiny{x} & \tiny{x} & \tiny{x} & \tiny{x} & \tiny{x} & \tiny{mod} & \tiny{x} & \tiny{4} & \tiny{guest} \\
\hline
\tiny{x} & \tiny{x} & \tiny{east} & \tiny{exist} & \tiny{exist} & \tiny{x} & \tiny{x} & \tiny{x} & \tiny{x} & \tiny{x} & \tiny{x} & \tiny{cheap} & \tiny{x} & \tiny{4} & \tiny{guest} \\
\hline
\tiny{x} & \tiny{x} & \tiny{east} & \tiny{exist} & \tiny{exist} & \tiny{x} & \tiny{x} & \tiny{x} & \textbf{\tiny{x}} & \tiny{x} & \tiny{x} & \tiny{cheap} & \tiny{x} & \tiny{4} & \textbf{\tiny{guest}} \\
\hline
\tiny{x} & \tiny{x} & \tiny{east} & \tiny{exist} & \tiny{exist} & \tiny{x} & \tiny{x} & \tiny{x} & \tiny{x} & \tiny{x} & \tiny{x} & \tiny{cheap} & \tiny{x} & \tiny{4} & \tiny{guest} \\
\hline
\end{tabularx}
} 
Please review the table. Would you like to reject any of the retrieved items?It looks like we've reached the end of our dialogue. If you have any more questions or need further assistance in the future, feel free to reach out. Have a great day!\\
\hline
\end{tabularx}
\caption{Instance of Confusion in a Conversation}
\label{xconfusion}
\end{figure}
In the conversation illustrated in Figure \ref{inconsist}, there is an inconsistency between the indicated configurations for predefined slots in the labeled goal of MultiWOZ and the utterances in the conversation. The user initially inputs "Cambridge to Bishops Stortford" as the departure and destination, in order. While, in another utterance in the middle of the conversation, the user indicates, "I need to depart from Bishops Stortford and go to Cambridge," in this way changes the configurations of departure and destination. In contrast, the MultiWOZ labeled goal does not portray this change, and incorrectly maintains "Cambridge" as the departure and "Bishops Stortford" as the destination, contrary to the user's changed preferences in the conversation. In this conversation, there is no issue with the implemented TOD; consequently it works without any changes. While to evaluate conversation completion, the information of the presented entity must be compared with the correct goal in the conversation, and not an incorrectly labeled goal.\\
\begin{figure}[H]
\small
\begin{tabularx}{\textwidth}{|p{0.6cm}|X|}
\hline
TOD & enter query \\
\hline

User & Can I get a train \textbf{from Cambridge to Bishops Stortford}? \\
\hline

TOD & \parbox[t]{\hsize}{The current values are: departure: Cambridge, destination: Bishops Stortford. Are there any other constraints besides the ones already mentioned?
} \\
\hline

User & On Thursday after 20:15. \\
\hline

TOD & \parbox[t]{\hsize}{I am updating your user preferences based on the constraints you have entered. Please hold on for a moment while I make these changes.

 \{ 'departure': 'cambridge', 'day': 'thursday', 'leaveat': '20:15', 'destination': 'bishops stortford'\} 
 Is there any other constraint besides the ones already mentioned?
} \\
\hline
User & No. I need to depart \textbf{from Bishops Stortford and go to Cambridge} on Thursday after 20:15. \\
\hline
TOD & \parbox[t]{\hsize}{Sure, I am updating your user preferences based on the constraints you've entered. Just a moment! You'd like to depart from Bishops Stortford and go to Cambridge on Thursday after 20:15 by train. Is there any other constraint besides the ones already mentioned?
} \\
\hline
User & Yes, that train would work better for me. Can you book tickets for 7 people please? \\
\hline
TOD & \parbox[t]{\hsize}{Updating your preferences based on the provided constraints. Please wait a moment while I make the necessary adjustments.
 'departure': 'bishops stortford', 'day': 'thursday', 'bookpeople': '7', 'leaveat': '20:15', 'destination': 'cambridge'. Is there any other constraint besides the ones already mentioned?
} \\
\hline
User & Great! Thank you! \\
\hline
TOD & It seems there might be some missing or unclear information in the provided text. NaN typically stands for "Not a Number," which means some data might not have been properly retrieved. 
To help clarify, here is the provided information in a table format:
\parbox[t]{\hsize}{
\begin{center}
\begin{tabular}{c|c|c|c|c|c|c}
\hline
\tiny{date} & \tiny{day} & \tiny{from} & \tiny{to} & \tiny{duration} & \tiny{time} & \tiny{price} \\
\hline
\tiny{x} & \tiny{Thursday} & \tiny{Bishops Stortford} & \tiny{Cambridge} & \tiny{x} & \tiny{x} & \tiny{x} \\
\hline
\end{tabular}
\end{center}
}
Please review the table and let me know if the retrieved items are correct or if any corrections are needed. Would you like to accept or reject these items?
 That concludes our conversation. If you have any more questions in the future, feel free to ask. Have a great day! \\
\hline

\multicolumn{2}{|X|}{%
\parbox[t]{\hsize}{
Extracted\_info: \{ \textbf{'departure':'bishops stortford'}, 'day': 'thursday', 'bookpeople': '7', 'leaveat': '20:15', \textbf{'destination': 'cambridge'} \} \newline
MultiWOZ\_info: \{'bookpeople': ['7'], 'day': ['thursday'], \textbf{'departure': ['cambridge']}, \textbf{'destination': ['bishops stortford']}, 'leaveat': ['20:15']\}}
} \\
\hline

\end{tabularx}

\caption{Inconsistent Conversation}
\label{inconsist}
\end{figure}
In certain MultiWOZ conversations with issues, such as confusion between the domain caption and configuration of a predefined slot type, and inconsistency in the MultiWOZ labeled goal, the problem in MultiWOZ user utterances that causes these issues is identified and the way to overcame these obstacles explained. While the Figure \ref{xcompleted} illustrates an example of the completed conversation. 

\begin{figure}[H]

\small

\begin{tabularx}{\textwidth}{|p{0.6cm}|X|}

\hline

TOD & enter query \\
\hline

User & Hello. Can you suggest a French restaurant in the north end? \\
\hline

TOD & \parbox[t]{\hsize}{
The values are:  area: north, food: french

Is there any other constraint besides the ones already mentioned?
} \\
\hline
User & I am interested in the one in the north. Could I have their postcode and address? \\
\hline
TOD & \parbox[t]{\hsize}{
I'm updating your user preferences based on the constraints you've entered. Please hold on for a moment while I make the necessary changes. Thank you for your patience!
The values of your request are: area: North, food: French
You are interested in a French restaurant in the North and specifically would like their postcode and address.\\
Are there any other constraints besides these?
} \\
\hline

User & Yes, that will be all. Thanks. \\
\hline

TOD & Certainly! Here is the information you provided in the form of a table:
\newline
\parbox[t]{\hsize}{
\begin{tabularx}{\hsize}{c|c|c|c|c|c|c|c|c}
\hline
\tiny{ID} & \tiny{Addr} & \tiny{Area} & \tiny{Cuisine} & \tiny{Rating} & \tiny{Type} & \tiny{Num} & \tiny{Phone} & \tiny{Postcode} \\
\hline
\tiny{x} & \tiny{x} & \tiny{north} & \tiny{french} & \tiny{x} & \tiny{restaurant} & \tiny{x} & \tiny{x} & \tiny{x} \\
\hline
\end{tabularx}
} 
Please indicate if you reject the retrieved items or not. 
 It appears that we've reached the end of our dialogue. If you have any more questions in the future, feel free to ask. Have a great day! \\
\hline

\multicolumn{2}{|X|}{%
\parbox[t]{\hsize}{
Extracted\_info: \{  'area': 'north', 'food': 'french' \} \newline
MultiWOZ\_info: \{ 'area': ['north'], 'food': ['french'], 'name': ['two two'], 'pricerange': ['expensive'] \}
}
} \\
\hline
\end{tabularx}
\caption{Completed Conversation}
\label{xcompleted}
\end{figure}
\subsection{Performance Examination }
\label{Performance Examination}
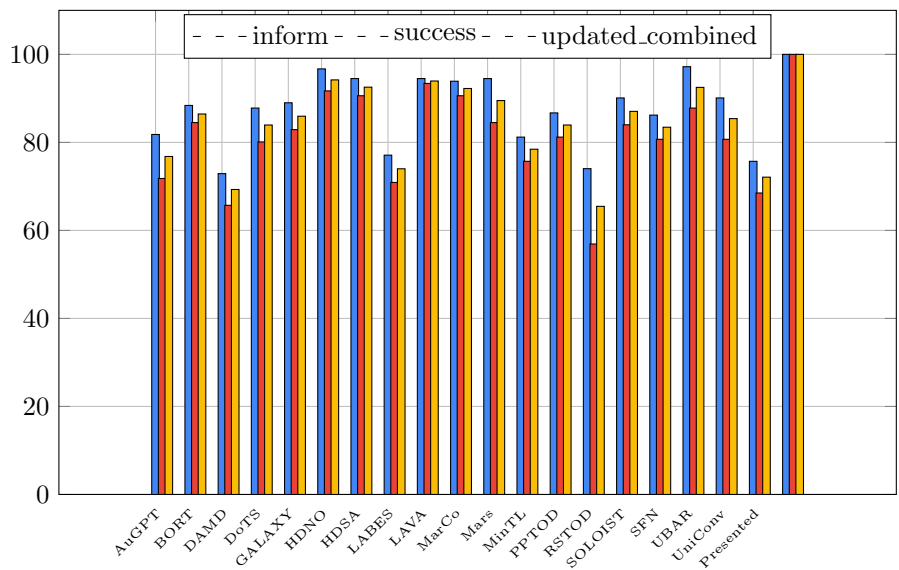
\begin{figure}[H]

    \centering
    \begin{tikzpicture}
        \begin{axis}[
            width=12.6cm,
            height=8cm,
            xlabel={},
            ylabel={},
            title={},
            xtick={0,1,...,18},  
            xticklabels={
                AuGPT, BORT, DAMD, DoTS, GALAXY, HDNO, 
                HDSA, LABES, LAVA, MarCo, Mars, MinTL, 
                PPTOD, RSTOD, SOLOIST, SFN, UBAR, 
                UniConv, Presented
            },
            xticklabel style={font=\tiny ,rotate=45, anchor=north east},
            ymin=0, ymax=110,
            bar width=0.15cm, 
            legend style={at={(0.5,0.90)}, anchor=south, legend columns=-1},
            grid=major,
            enlarge x limits=0.15, 
        ]

        \addplot[
            ybar,
            bar width=0.10cm,
            fill=xblue,
            draw=black,
        ] coordinates {
            (0,81.8)(1,88.4)(2,72.9)(3,87.8)(4,89.0)(5,96.7)
            (6,94.5)(7,77.1)(8,94.5)(9,93.9)(10,94.5)(11,81.2)
            (12,86.7)(13,74.03)(14,90.1)(15,86.2)(16,97.2)(17,90.1)(18,75.7)(19,100)
        };

        \addplot[
            ybar,
            bar width=0.10cm,
            fill=xred,
            draw=black,
        ] coordinates {
            (0+0.2,71.8)(1+0.2,84.5)(2+0.2,65.7)(3+0.2,80.1)(4+0.2,82.9)(5+0.2,91.7)
            (6+0.2,90.6)(7+0.2,70.9)(8+0.2,93.4)(9+0.2,90.6)(10+0.2,84.5)(11+0.2,75.7)
            (12+0.2,81.2)(13+0.2,56.91)(14+0.2,84.0)(15+0.2,80.7)(16+0.2,87.8)(17+0.2,80.7)(18+0.2,68.5)(19+0.2,100)
        };

        \addplot[
            ybar,
            bar width=0.10cm,
            fill=xyellow,
            draw=black,
        ] coordinates {
            (0+0.4,76.8)(1+0.4,86.45)(2+0.4,69.3)(3+0.4,83.95)(4+0.4,85.95)(5+0.4,94.2)
            (6+0.4,92.55)(7+0.4,74.0)(8+0.4,93.95)(9+0.4,92.25)(10+0.4,89.5)(11+0.4,78.45)
            (12+0.4,83.95)(13+0.4,65.47)(14+0.4,87.05)(15+0.4,83.45)(16+0.4,92.5)(17+0.4,85.4)(18+0.4,72.1)(19+0.4,100)
        };

        \legend{inform, success, updated\_combined}
        \end{axis}
    \end{tikzpicture}
    \caption{Comparing the Performance of Priors and Created TODs }
    \label{x_comparison}
\end{figure}

Paper \citep{nekvinda2021shades} concentrates on the challenge of comparing TODs exploiting MultiWOZ dataset. They implement the common evaluation metrics for TODs, with the input parameter being the predictions made by the investigated TOD for conversations in the test dataset of MultiWOZ. Predictions of previously created TODs are in the corresponding \href{https://github.com/Tomiinek/MultiWOZ_Evaluation/tree/master/predictions} {GitHub repository}. \\
While in this paper, the predictions corresponding to conversations that are in more than a domain, which are out of this paper's problem, are eleminated from the predictions of previously created TODs. In this way, the predictions of the investigated TODs in the test dataset were collected. Then, the evaluation method implemented by \citep{nekvinda2021shades} is performed with these predictions as input to calculate the inform and success for previously created TODs in the test dataset.\\
The other metric for evaluating TODs is 'combined,' which integrates inform, success and BLEU. This is commonly calculated by cumulating the mean of inform and success, to the BLEU. While the concentration in this paper is on task completion, consequently   the BLEU score is excluded from this combination. Instead, the mean of the inform and success is calculated, indicated as 'updated\_combined,' to compare the performance with prior TODs in task completion. Table \ref{xcompare} and diagram \ref{x_comparison} illustrate the performance of presented model in this paper and previously created TODs.
\begin{table}[]
\caption{Comparing the Performance of Previous TODs with the Created Model}

\begin{tabular}{cccc}
\toprule
\textbf{model} & \textbf{inform} & \textbf{success} & \textbf{updated\_combined} \\
\midrule
AuGPT\citep{kulhanek2021augpt}   & 81.8  & 71.8 & 76.8 \\
BORT  \citep{sun2022bort}    & 88.4  & 84.5 & 86.45 \\
DAMD\citep{zhang2020task}    & 72.9  & 65.7 & 69.3 \\
DoTS \citep{jeon2021domain}   & 87.8  & 80.1 & 83.95 \\
GALAXY\citep{he2022galaxy}  & 89.0  & 82.9 & 85.95 \\
HDNO \citep{wang2020modelling}   & 96.7  & 91.7 & 94.2 \\
HDSA \citep{chen2019semantically}   & 94.5  & 90.6 & 92.55 \\
LABES \citep{zhang2020probabilistic}  & 77.1 & 70.9 & 74.0 \\
LAVA \citep{lubis2020lava}   & 94.5  & 93.4 & 93.95 \\
MarCo \citep{wang2020multi}   &  93.9  & 90.6 & 92.25 \\
Mars \citep{sun2022mars}     & 94.5  & 84.5 & 89.5 \\
MinTL \citep{lin2020mintl}   & 81.2  & 75.7 & 78.45 \\
PPTOD \citep{su2022multi}  & 86.7  & 81.2 & 83.95 \\
ProTOD \citep{xprotod}  &  74.03  &  56.91 &  65.47 \\
RSTOD \citep{cholakov2022efficient}  & 90.1  & 84.0 & 87.05 \\
SOLOIST \citep{peng2021soloist} & 86.2  & 80.7 & 83.45 \\
SFN \citep{mehri2019structured}    & 97.2  &87.8  & 92.5 \\
UBAR \citep{yang2021ubar}   & 90.1  & 80.7  & 85.4 \\
UniConv \citep{le2020uniconv} & 75.7 & 68.5 & 72.1 \\
Presented Model & 100   & 100   & 100 \\
\botrule
\end{tabular}
\label{xcompare}
\end{table}

Considering the explanation of the presented model's capabilities in management, even in certain problematic conversations in MultiWOZ 2.2, in whole conversations within the test dataset, the implemented TOD provides complete information of entities that is congruent with the user preferences. In other words, not only are the entities centered on user preferences presented to the user, and there are no contradictions between the characteristics of the presented entities with the correct current user goal and the information state of the conversation, but also the whole information of entire congruent entities to preferences in predefined slots is presented to the user in an order indicated centered on the text\_part of preferences exploiting entity\_ranker. This mechanism ensures that there is no drop in success compared to inform. In other words, if the predefined slots in a conversation are correctly identified, while the preferences beyond predefined slots considered in indicating the order of presented entitles, no information that must be conveyed to the user is omitted in the presented entities. Without contradicting this point, the evaluation of the presented method indicates not only complete inform but also complete success.\\
Comparing the inform of the presented method with those of previous works exposes an improvement of at least 2.8 points over the maximum inform from prior models (SFN) on the test dataset. In Figure \ref{x_comparison}, it is evident that the success of previous methods is consistently lower than their inform. Table \ref{xcompare} exhibits that the gap between inform and success is different across models, ranging from about 1 point in the case of LAVA to 10 points in AuGPT and Mars. The ProTOD is based on in-context learning of LLMs. To improve proactivity, collecting entity information in ProTOD is not limited to the database. In calculating inform in the ProTOD paper and consequently in this paper, the presumption is that passing entities without contradiction with preferences, even if the entity is not in the database, improves inform. While this presumption to calculate inform in TODs is challenging. \\
The presented method improves the updated\_combined metric (which combines inform and success), with an increase of circa 5.8 points over the maximum updated\_combined of previous models (HDNO). The updated\_combined of LAVA is comparable with HDNO, with a difference of about 6 points compared to the presented method. It must be considered that the gap between inform and success of LAVA is the smallest, while its success is the maximum in the investigated previous methods in the test dataset. In contrast, updated\_combined of SFN, which its inform is maximum, is in a lower position in investigated previous works due to a drop in success.\\
Compared to the performance of previous works, the presented method improves inform by 2.8 points, success by 6.6 points, and updated\_combined by 5.8 points, with no gap between inform and success. It is important to point out that the maximum performances of the previous models correspond to different TODs—no individual model in prior works excels in whole metrics. In a way that, in investigated previous works, the model with the maximum inform is not essentially the model with the maximum success.

\section{Conclusion}
\label{Conclusion}
In this paper, we created a model for collaborative and proactive management of task-oriented conversations. Considering the limitations of the models from previous works, the presented model conceptualized centered on the information state approach to dialogue management. This model incorporates intermediate informational components that are crucial for the proactive management of task-oriented conversations. User preferences are modeled as the predefined slot and text\_part informational components. By examining challenging conditions in task-oriented conversations, we identified other informational components whose exploitation could mitigate error-prone conditions. These informational components leads to limited possible information states. Consequently, dialogue moves created, and the update strategy constructed. Inspired by task decomposition and external planner-aided planning methods, the model is implemented centered on in-context learning of LLMs. The whole domain-dependent information is extracted in pre-processing, creating the constant portion of the information state, and is exploited as input parameters for procedures. With this consideration, the presented model is independent of the domain. The exploited entity retrieval mechanism enables presenting the whole congruent entities to user preferences in an orderly manner. The conducted experiments leveraging the conversations of MultiWOZ test, without changing domain in a conversation, indicated zero failures in task completion (i.e., maximal inform and success), illustrating the effectiveness of the presented model in managing task-oriented conversations in different domains. Compared to previously presented models, this model considerably improves inform and success task completion metrics. It is possible to create an instruction collection exploiting the conversations with the created TOD, which will include intermediate information to create an instruction-tuned LLM as a proactive, collaborative TOD.
\section*{Competing interests}
 The authors declare no competing interests.
\section*{Authors' contributions }
Arezoo Saedi: Conceptualization, Investigation, Methodology, Project administration, Software, Validation, Visualization, Writing – original draft\\
Afsaneh Fatemi, Mohammad Ali Nematbakhsh, Sophie Rosset, Anne Vilnat: Conceptualization, Methodology, Project administration, Resources, Supervision, Validation, Writing – review and editing
\section*{Declaration of generative AI and AI-assisted technologies in the writing process}
During the preparation of this work the author(s) used [ChatGPT] in order to [editing]. After using this tool/service, the author(s) reviewed and edited the content as needed and take(s) full responsibility for the content of the publication.
\section*{Acknowledgments}
This research was conducted without any external funding.

\end{document}